\documentclass[conference]{IEEEtran}
\IEEEoverridecommandlockouts
\usepackage{cite}
\usepackage{amsmath,amssymb,amsfonts}
\usepackage{algorithmic}
\usepackage{graphicx}
\usepackage{textcomp}
\usepackage{xcolor}

\usepackage{sistyle}
\SIthousandsep{,}
\usepackage{graphicx}
\usepackage{url}
\usepackage{nohyperref}
\usepackage{bm}

\def\BibTeX{{\rm B\kern-.05em{\sc i\kern-.025em b}\kern-.08em
    T\kern-.1667em\lower.7ex\hbox{E}\kern-.125emX}}
\begin{document}

\title{A light neural network for modulation detection under impairments\\
}

\author{\IEEEauthorblockN{Thomas Courtat}
\IEEEauthorblockA{\textit{Thales SIX Theresis} \\
1 av. Augustin Fresnel, 91120 Palaiseau, France \\
thomas.courtat'at'thalesgroup.com
}
\IEEEauthorblockN{Hélion du Mas des Bourboux}
\IEEEauthorblockA{\textit{Thales SIX Theresis} \\
1 av. Augustin Fresnel, 91120 Palaiseau, France \\
helion.dumasdesbourboux'at'thalesgroup.com
}
}

\IEEEoverridecommandlockouts
\IEEEpubid{\makebox[\columnwidth]{978-0-7381-1316-6/21/\$31.00~\copyright2021 IEEE \hfill} \hspace{\columnsep}\makebox[\columnwidth]{ }}

\maketitle

\IEEEpubidadjcol

\begin{abstract}
We present a neural network architecture able to efficiently detect
modulation scheme in a portion of I/Q signals.
This network is lighter by up to two orders of magnitude
than other state-of-the-art  architectures working on the same or similar tasks.
Moreover, the number of parameters does not depend on the signal duration,
which allows processing stream of data, and results in a signal-length invariant
network.
In addition, we have generated a dataset based on the simulation of
impairments that the propagation channel and the
demodulator can bring to recorded I/Q signals:
random phase shifts, delays, roll-off, sampling rates, and frequency offsets.
We benefit from this dataset to train
our neural network
to be invariant to impairments and quantify its accuracy at
disentangling between modulations under realistic real-life conditions.
Data and code to reproduce the results are made publicly
available.
\end{abstract}

\begin{IEEEkeywords}
Machine learning, deep learning, modulation recognition, propagation channel, data augmentation
\end{IEEEkeywords}

\section{Introduction}

During the last few years, a lot of effort has been put into applying
the performances of machine learning to the physical layer of radio
transmission. Toward this goal, multiple directions are investigated:
of interest in this study is
modulation classification through supervised learning
\cite{10.1007/978-3-319-44188-7_16}.

In machine learning terminology, modulation classification is the task consisting in recognizing
what modulation has been used to produce the received noisy and impaired radio signal.
Such a classification would give insights on the nature of the different emitters present
in the radio spectrum. Dynamic Spectrum Access (DSA) protocols would benefit
from such insights and allow better adaptation to dynamic and diverse environments.

A large step towards machine learning for telecommunication signal classification
was accomplished by~\cite{10.1007/978-3-319-44188-7_16}
through the publication of a public dataset for radio modulation classification,
along with an artificial neural network architecture.
Following this release multiple publications presented neural networks
and analyses with respect to this dataset, e.g. \cite{8737463,
8449065, 2019arXiv190105850R,
shi2019deep,
2019arXiv191104970T, 8936957, 2020arXiv200101395T}.
\cite{2018ISTSP..12..168O} showed that machine learning (ML) based modulation
classifiers already outperform traditional techniques based on
higher order statistics. On the other hand \cite{8645089} showed that
even though ML based classifiers give better results, they can be less robust
to data with impairments not present in the training set.
This outlines the need to feed realistic and complete datasets to
machine learning algorithms.

This study presents a novel
neural network architecture that outperforms existing ones in the modulation
recognition task.
It is lighter than previously published
networks~\cite{10.1007/978-3-319-44188-7_16,2018ISTSP..12..168O} and is built to be
invariant under signal duration.
We also develop a synthetic dataset generator that allows to better
control the sets of impairments and better understand their effects on
the accuracy of our classifier.

This article is organized as follows. Section~\ref{section::data_sets}
presents all the datasets, either publicly available or developed here.
Then, section~\ref{section::neural_network_architectures} details the
three state of the art architectures and the two developed for this study.
They are compared in section~\ref{section::comparing_the_different_architectures}
with respect to their accuracy at classifying modulations and we show that
our network outperforms the others, while being two to ten times lighter.
We study the performances of our architecture under variation of the
signal length, and under frequency shifts
in section~\ref{section::performances_of_Mod_LRCNN}.
We present a conclusion in section~\ref{section:conclusion}.

The dataset generated for this article and the Python/TensorFlow \cite{tensorflow2015-whitepaper}
implementation of deep learning architectures under study are made publicly available.

\section{datasets}
\label{section::data_sets}

The industry standard dataset,
and its following updates, for modulation classification in radio is given
by \cite{grcon, 10.1007/978-3-319-44188-7_16, o2016unsupervised}.
The first release, \texttt{RadioML2016.04C} (ML: Machine Learning), is composed of $11$~modulations:
8PSK, \mbox{AM-DSB}, \mbox{AM-SSB}, BPSK, CPFSK, GFSK, PAM4, QAM16, QAM64, QPSK, and WBFM,
with $20$~evenly spaced bins in signal-to-noise ratio (SNR), ranging from $-20$ to $18$~dB.
The set is composed of $\num{162060}$~examples,
consisting in $128$~samples of I/Q (in-phase/quadrature) signals.
The simulated synthetic data were produced using software defined radio programmed with
GNU radio \cite{Blossom:2004:GRT:993247.993251}.

Three releases have expanded and completed the set.
\texttt{RadioML2016.10A} expanded to $\num{220000}$ the number of examples.
\texttt{RadioML2016.10B} provides $\num{1200000}$~examples on the same
grid of SNR, but removes the \mbox{AM-SSB} modulation, leading to a total of
$10$~classes.
\texttt{RadioML2018.01A} \cite{2018ISTSP..12..168O} provides a total of
$\num{2555904}$~examples, $1024$~samples long,
with a signal-to-noise ratio ranging from $-20$ up to $30$~dB.
Along with synthetic data, this set provides radio signals propagated
through real indoor environment, transmitted and received via two
universal software radio peripherals (USRP).
This former dataset expands to $24$ the number of different
modulations classes:
128APSK, 128QAM, 16APSK, 16PSK, 16QAM, 256QAM, 32APSK, 32PSK, 32QAM,
4ASK, 64APSK, 64QAM, 8ASK, 8PSK,
\mbox{AM-DSB-SC}, \mbox{AM-DSB-WC}, \mbox{AM-SSB-SC}, \mbox{AM-SSB-WC},
BPSK, FM, GMSK, OOK, OQPSK, and QPSK.
For simplicity, in this study we limit the datasets to positive SNR,
we verify nonetheless that similar results are obtained on the whole range.
All of these datasets are publicly
available\footnote{\url{https://www.deepsig.io/datasets}}.

In order to independently study the performances of machine learning in
modulation classification, we develop a synthetic custom dataset.
The received signals are simulated through a baseband equivalent model.
In addition, this module allows us to tune different parameters which
are fixed or unknown in the previously defined datasets. As a consequence it allows us to study
its robustness against parameters, while improving upon it.
Seven linear modulations are simulated:
BPSK, PSK8, QAM16, QAM32, QAM64, QAM8, and QPSK, with $5$ evenly spaced bins of SNR
from $0$ to $40$~dB. We generate $\num{175000}$~examples, i.e. $\num{5000}$
per (SNR, modulation) pairs. The I/Q signal is produced for $1024$~samples.
A vast range of impairments brought by the propagation channel and the demodulator
are added to the baseline dataset:
random phase shifts, delays, roll-off, sampling rates, and additive Gaussian noise.
We also produce an additional dataset enhanced with an extra impairment:
relative frequency offsets.
This allows us to better study its individual impact on modulation
classification.
Hereafter, we refer to this dataset as \texttt{AugMod},
for ``augmented modulation'' dataset.
The range of the parameters are given in table~\ref{table::set_of_impairments}.
The dataset is publicly available\footnote{\url{https://augmod.blob.core.windows.net/augmod/augmod.zip}}.
\begin{table}
    \centering
    \begin{tabular}{ll}
    Impairment & Range\\
    \noalign{\vskip 1mm} 
    \hline
    \hline
    \noalign{\vskip 2mm}
    $T_{\mathrm{sample}}/T_{\mathrm{symbol}}$ & $\left[ 0.3,0.5 \right]$ \\
    Phase & $\left[ 0, 2\pi \right]$ \\
    Delay &  $\left[ 0,1 \right]$ \\
    Roll off & $\left[ 0.1,0.5 \right]$ \\
    $\mathrm{SNR}$ & $\left\{ 0, 10, 20, 30, 40\right\}$ \\
    Relative frequency offset, $\Delta f$ &
        $\pm [10^{-6}, 5\times10^{-1}]$ \\
    %    $\left\{-1,1\right\}*\left\{1,5\right\}*10^{-\left\{1,2,3,4,5,6\right\}}$ \\
    \noalign{\vskip 2mm}
    \end{tabular}
    \caption{Set of impairments simulated in the \texttt{AugMod}
    synthetic dataset, developed in this study.}
    \label{table::set_of_impairments}
\end{table}

As a result we benefit from five different datasets, with positive SNR,
with both synthetic and indoor-propagated signals,
to perform modulation classification under impairments.
The first four datasets are public:
\texttt{RadioML2016.04C} has $\num{81030}$~examples,
\texttt{RadioML2016.10A} has $\num{110000}$~examples,
\texttt{RadioML2016.10B} has $\num{600000}$~examples,
and \texttt{RadioML2018.01A} has $\num{1572864}$~examples.
The fifth dataset \texttt{AugMod} has been generated for this work and contains $\num{175000}$~examples.
Each dataset is split into two halves, one for training and the other
for testing. Each individual signal is normalized by its root mean
square, to have a power of $1$.

\section{Neural network architectures}
\label{section::neural_network_architectures}

Along with the available dataset, \cite{10.1007/978-3-319-44188-7_16} presents
a convolutional neural network (ConvNet, \cite{LeCun:1989:HDR:2969830.2969879})
performing modulation classification,
hereafter referred as ``RML-ConvNet'' (RML: Radio Machine Learning). This network
processes the complex I/Q signal as a two-dimensional image,
with a single ``color'' channel. As it is presented, this network has
$\num{2829399}$~parameters, when the I/Q signal has $128$~samples
and the dataset has $7$~different classes.
The architecture is not invariant with the number of samples;
this imposes to train a different network for every possible
length of the input signal. Furthermore, a signal given with 1024 samples
would multiply the number of parameters by approximately one order of
magnitude, compared to the one for $128$.
This aspect produces a hardly scalable architecture for longer
signals.
Table~\ref{table::nb_parameters_networks} gives for two different
length of signals, $128$ and $1024$, the number of parameters, or weights, of the neural
networks studied here.
\begin{table*}
    \centering
    \begin{tabular}{l|lll|ll}
    Number of & RML-ConvNet & RML-CNN/VGG & RML-ResNet & Mod-LCNN & Mod-LRCNN\\
    %& ConvNet & CNN/VGG & ResNet & LCNN   & LRCNN\\
    signal samples &         &         &        & (ours) & (ours)\\
    \noalign{\vskip 1mm} 
    \hline
    \hline
    \noalign{\vskip 2mm}
    $128$  & $\num{2829399}$  & $\num{199111}$ & $\num{179303}$ & $\num{37487}$ & $\num{97663}$ \\
    $1024$ & $\num{21179479}$ & $\num{256455}$ & $\num{236647}$ & $\num{37487}$ & $\num{97663}$  \\
    \noalign{\vskip 2mm}
    \end{tabular}
    \caption{Number of parameters of the five different networks for signals
    with $128$ or $1024$~samples. For this table we choose $7$ output classes,
    slightly different number of classes do not yield significant changes in the order of magnitude
    of the number of parameters.}
    \label{table::nb_parameters_networks}
\end{table*}

In a more recent release of their work, \cite{2018ISTSP..12..168O}
presented an updated dataset, \texttt{RadioML2018.01A},
with $1024$~samples long signals.
They also developed two extra neural networks: ``RML-CNN/VGG''
and ``RML-ResNet''. The first network builds upon the already developed
RML-ConvNet network, but limits the explosion of the number of parameters
at $1024$~samples through a VGG network (Visual Geometry Group, \cite{Simonyan15}).
It is modified to fit a 1-dimensional
convolutional neural network (CNN).
The second network has a residual architecture (ResNet, \cite{7780459}).
ResNet has historically been invented to
be easier to train for deep neural networks.
Although both of these networks have less parameters than RML-ConvNet, as shown on
table~\ref{table::nb_parameters_networks}, they still suffer from
the augmentation of the number of parameters with the signal length.
For example, going from $128$ to $1024$~samples adds $30\%$ more
parameters.
Because of this aspect, they lack the ability to adapt to signals
of different sizes and, as for RML-ConvNet, must be re-trained
for each signal length.

We propose a lighter convolutional neural network to perform
modulation classification, invariant of the input signal length:
Light Modulation Convolutional Neural Network, ``Mod-LCNN''.
The complex I/Q signal is treated as
a one-dimensional signal with two channels.
These channels are expanded to higher dimension space through
consecutive 1-dimensional convolutional layers. Then through an average pooling
layer, the time dimension is collapsed to produce a one-dimensional
layer of dimension that of the last convolutional layer,
which is fed into a fully connected layer, and
a softmax~\cite{NIPS1989_195} layer to perform classification.
Each convolutional layer of kernel size~$7$, along with the first fully
connected layer, are followed by the rectified linear unit (ReLU, \cite{pmlr-v15-glorot11a})
activation function.
During training, we apply dropout~\cite{JMLR:v15:srivastava14a}
to the output weights of the first
fully connected layer, thus preventing overfitting.

\begin{figure*}[ht]
    \centering
    \includegraphics[width=0.58\linewidth]{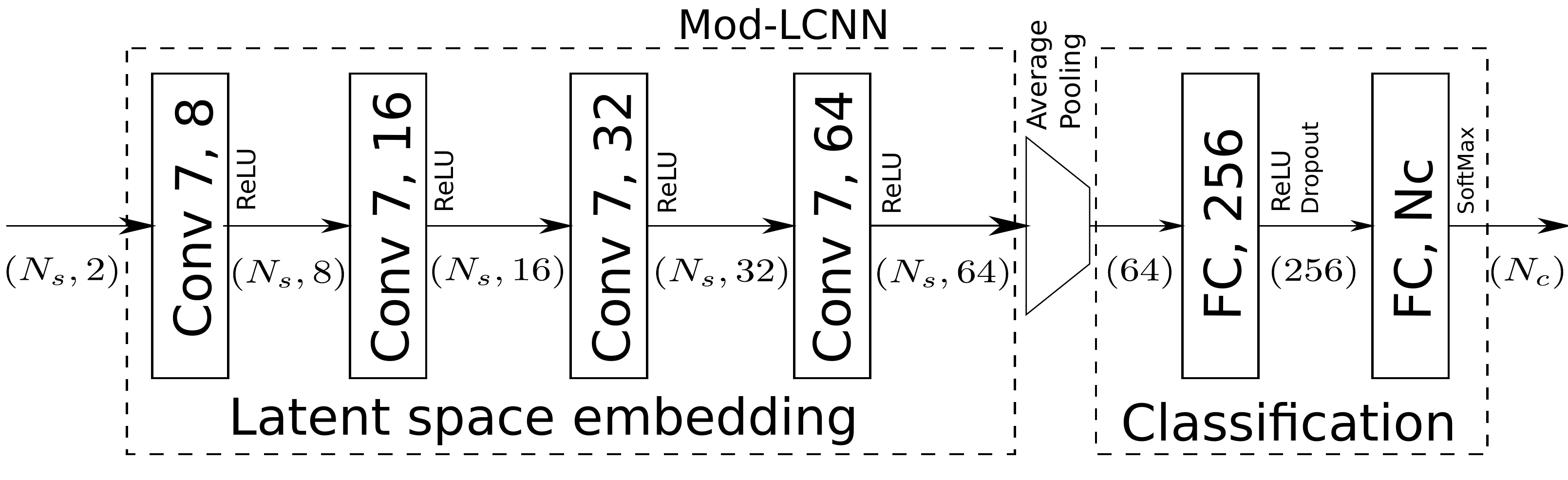}\\
    \includegraphics[width=0.98\linewidth]{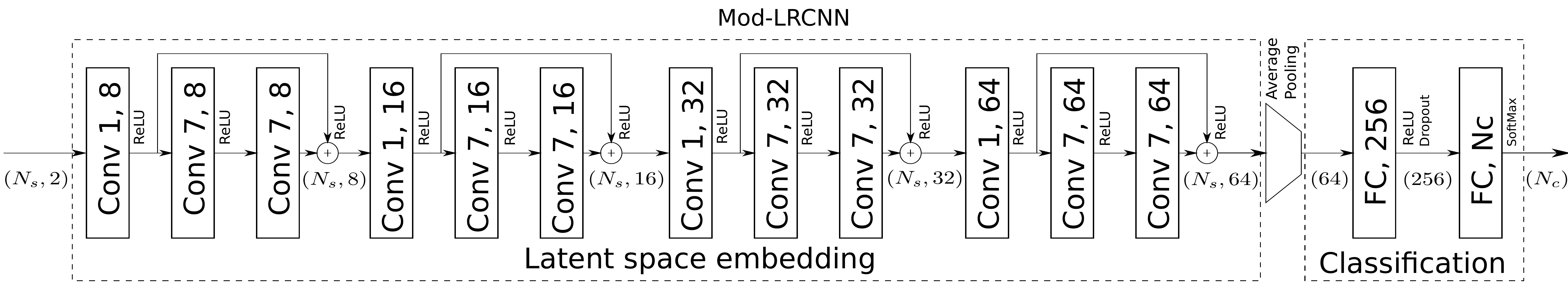}
    \caption{Architecture of Mod-LCNN (top) and Mod-LRCNN (bottom),
    the neural networks developed in this study.
    $N_{s}$ is the number of samples: $128$ or $1024$ in this study, $N_{c}$ is the number of
    output classes: $7$, $10$, $11$ or $24$ in this study.
    The 1-dimensional convolutions have a kernel size of $7$.
    During training the dropout rate is $0.5$.}
    \label{figure::design_custom_architecture}
\end{figure*}

We develop two different networks: ``Mod-LCNN'' and ``Mod-LRCNN''.
Both are presented on figure~\ref{figure::design_custom_architecture}.
These two networks have the structure presented above, they differ in
how each convolutional layer is applied. In the case of Mod-LCNN (top panel),
we use a regular CNN, Mod-LRCNN (bottom panel) is a ResNet~\cite{7780459}.
As a consequence each convolution step is split into three simple
convolutions. The first one has a kernel size $1$, allowing to expand
the filter dimension~\cite{2013arXiv1312.4400L}, the two following 
convolutions have a kernel size of $7$. The output of these last two
consecutive convolutions is added to the output of the first one, through
a skip connection (see figure~\ref{figure::design_custom_architecture}).

For these two networks, the number of parameters does not depend on the
signal duration. The consequence of this design is that the same
trained network can be used for signals of different lengths.
The resulting networks have $\num{37487}$~parameters for Mod-LCNN
and $\num{97663}$ for Mod-LRCNN (table~\ref{table::nb_parameters_networks}).
As shown on figure~\ref{figure::design_custom_architecture}, these two networks
can be modeled as two blocks.
The first one is a ``latent space embedding'', i.e. it extracts latent
features of the signal, invariant of its length. The second block is
a fully connected network that performs the ``classification''.
The average pooling layer serves thus as a bottleneck between these two
blocks.

Notice that these convolutional architectures can be seen as a particular class 
of recurrent neural networks (RNN: \cite{118724, 6796188}) with a sparse structure 
which is particularly interesting for computational cost consideration.

\section{Comparison of the different architectures}
\label{section::comparing_the_different_architectures}

We benefit from the five different datasets presented in
section~\ref{section::data_sets} to train and compare the five artificial
neural networks of section~\ref{section::neural_network_architectures}.
RML-ConvNet implementation is publicly provided by the
author\footnote{\url{https://github.com/radioML/examples}},
in Keras \cite{chollet2015keras},
with TensorFlow backend~\cite{abadi2016tensorflow}, we thus use the
same framework for all the other network architectures.
Following the publicly available implementation of RML-ConvNet, we initialize
all weights using the ``Glorot'' uniform initializer~\cite{pmlr-v9-glorot10a}
for convolutional layers
and through ``He'' normal initializer~\cite{10.1109/ICCV.2015.123} for fully connected layers.
The training is ran on a NVIDIA~1080~Ti.

The neural network weights are learned using the training set through the
Adam optimizer~\cite{2014arXiv1412.6980K}, to minimize
the categorical cross-entropy loss function.
Among the five datasets, two have $1024$~samples long signals: 
\texttt{RadioML2018.01A} and \texttt{AugMod}.
We train the networks on these two datasets twice: once on the full
signal duration, and another time keeping only the first $128$~samples.
The training is performed for $200$~iterations, or epochs, through each dataset with a
batch size of $512$~examples. Because of computation time,
all networks are trained for only $50$~iterations for \texttt{RadioML2018.01A},
when using the full $1024$~samples long signals.
The neural network implementations and the code to train them is publicly
available\footnote{\url{https://github.com/ThalesGroup/pythagore-mod-reco/}}.

Table~\ref{table::perfs_accuracy} presents the accuracy on the test
sets for the five datasets, over the five different neural networks.
The accuracy, in percent, is given by the proportion of correctly assigned
modulations on the test set, after the end of training.
Boldface texts highlight the best results for each dataset.
All networks perform relatively equally well on \texttt{RadioML2016.04C} and on
\texttt{RadioML2016.10B}.
RML-ConvNet and RML-CNN/VGG do not manage to reach as good performances
as other networks on
other datasets. This is explained by the too large number of parameters
for the first network, resulting in overfitting the training set.
For the second network this is explained by the depth of the network,
preventing the gradient updates to efficiently propagate through the network.
\begin{table*}
    \centering
    \begin{tabular}{l|lll|ll}
    dataset & RML-ConvNet & RML-CNN/VGG & RML-ResNet & Mod-LCNN & Mod-LRCNN\\
    % & ConvNet & CNN/VGG & ResNet & LCNN & LRCNN\\
     &  &  &  & (ours) & (ours)\\
    \noalign{\vskip 1mm} 
    \hline
    \hline
    
    \noalign{\vskip 1mm}
    $128$ samples  & & & & & \\
    \noalign{\vskip 1mm}
    \hline
    \noalign{\vskip 2mm}

    \texttt{RadioML2016.04C}  & $93$ & $92$ & $94$ & $93$ & \bm{$94$} \\
    \texttt{RadioML2016.10A}  & $84$ & $83$ & $90$ & $89$ & \bm{$91$} \\
    \texttt{RadioML2016.10B}  & $88$ & $90$ & $92$ & $92$ & \bm{$92$} \\
    \texttt{RadioML2018.01A}  & $50$ & $69$ & $76$ & $68$ & \bm{$76$} \\
    \texttt{AugMod} (ours)    & $61$ & $56$ & $65$ & \bm{$75$} & $72$ \\
   
    \noalign{\vskip 1mm}
    \hline
    \noalign{\vskip 1mm}
    $1024$ samples  & & & & & \\
    \noalign{\vskip 1mm}
    \hline
    \noalign{\vskip 2mm}
    
    \texttt{RadioML2018.01A} & $61$ & $87$ & $88$ & $85$ & \bm{$89$} \\
    \texttt{AugMod} (ours)   & $68$ & $14$ & $78$ & \bm{$83$} & $82$ \\
    \noalign{\vskip 2mm}
    
    \end{tabular}
    \caption{Accuracy of the five different neural network architectures
    on the different datasets.
    The performances are given for a signal of size $128$ for all datasets,
    and for $1024$~samples when available.
    Boldface texts highlight the best results for each datasets.}
    \label{table::perfs_accuracy}
\end{table*}

We confirm the results noted by \cite{2018ISTSP..12..168O}, RML-ResNet
gives indeed the best performances over all the datasets, when compared
to RML-ConvNet and RML-CNN/VGG.
Mod-LCNN, developed in this study, outperforms or equals RML-ResNet when testing
on the \texttt{AugMod} dataset, however it fails at giving good results
on \texttt{RadioML2018.01A}. This can be interpreted by the too small
number of parameters. Adding more layers would reduce the performances
by producing a too deep architecture, harder to train.
Mod-LRCNN manages to outperforms or equals all the other networks, building on
Mod-LCNN performances, but adding a residual network
architecture.
Mod-LRCNN has similar performances than RML-ResNet on all \texttt{RadioML}
datasets, and outperforms it by up to $7$\% on the \texttt{AugMod} dataset.

\begin{figure}[ht]
    \centering
    \includegraphics[width=0.99\linewidth]{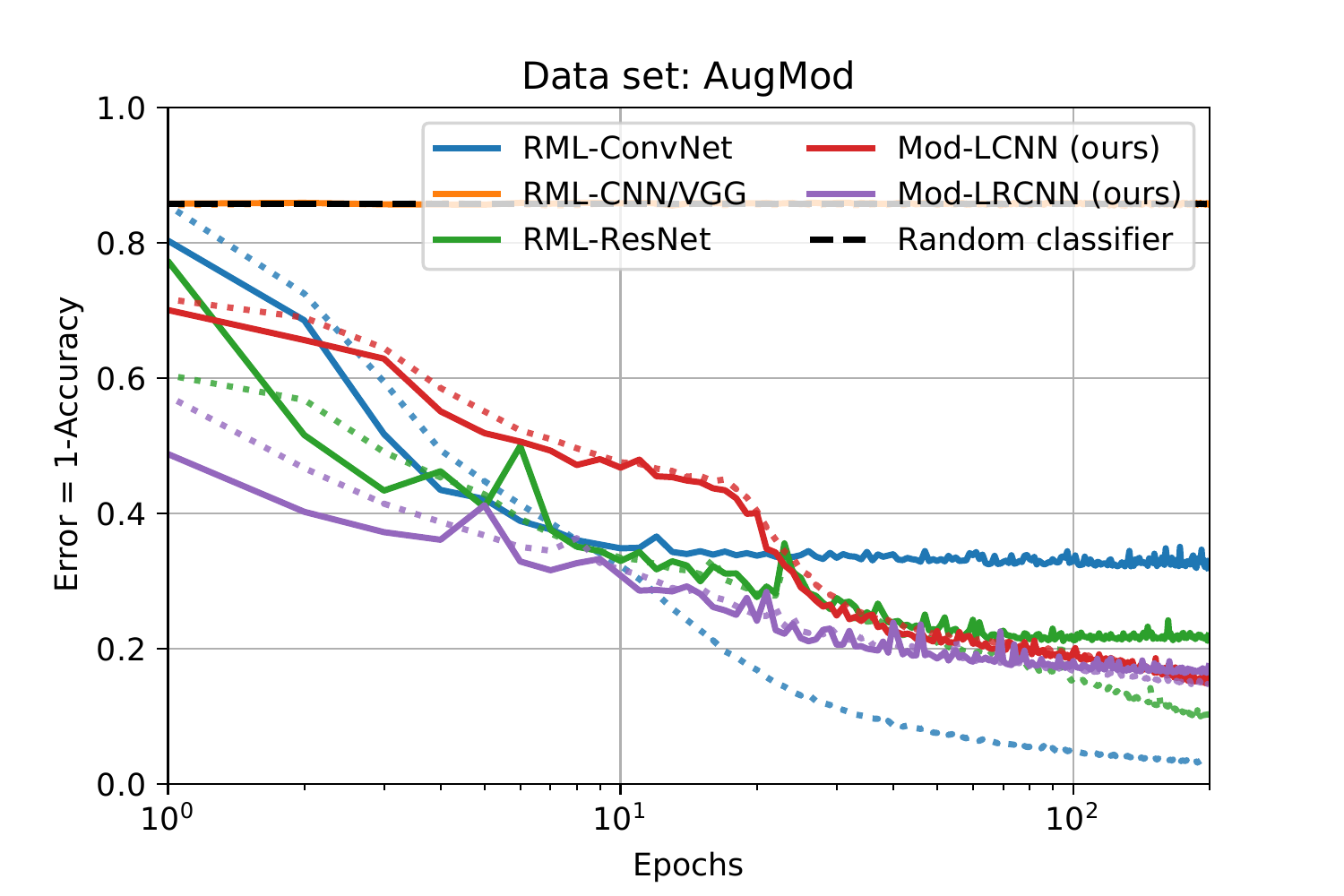}
    \caption{Error rate as a function of the number of epochs for the five different
    neural network architectures, compared with the performances of a
    random classifier.
    Solid curves are for the test set and dotted curves for the training set.
    The comparison is performed with the \texttt{AugMod} dataset on
    $1024$~samples long signals.}
    \label{figure::learning curves}
\end{figure}

Figure~\ref{figure::learning curves} presents the learning curves, i.e.
the error rate as a function of the number of epochs, for all the networks,
on the \texttt{AugMod} dataset, with $1024$~samples. Unbroken curves
give the results on the test sets, and dotted curves
on the training sets.
This figure outlines the advantages of the Mod-LRCNN architecture:
it outperforms other architectures with the lowest error rate, converges faster and
continuously, and is less prone to overfitting.

\begin{figure*}[ht]
    \centering
    \includegraphics[width=0.48\linewidth]{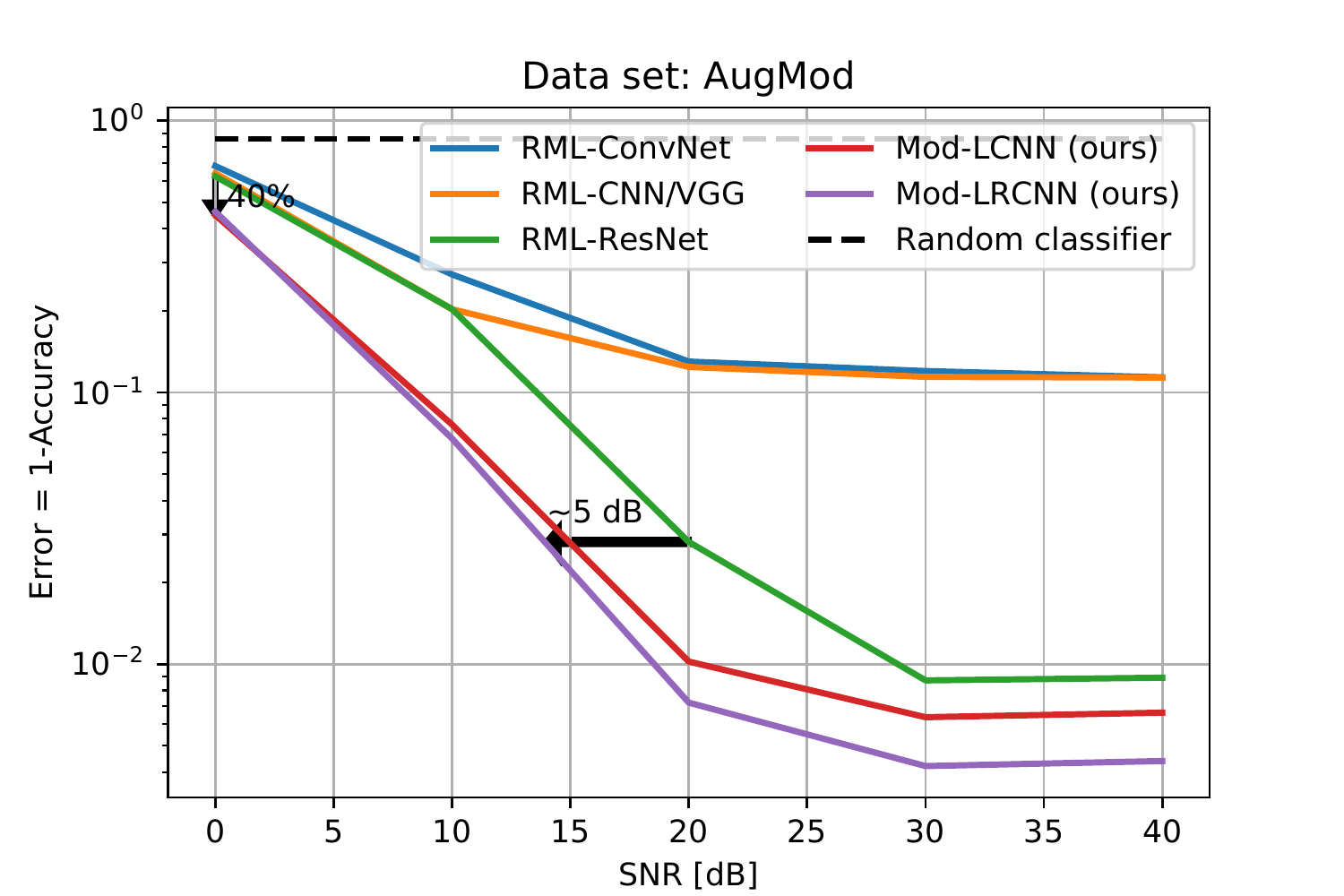}
    \includegraphics[width=0.48\linewidth]{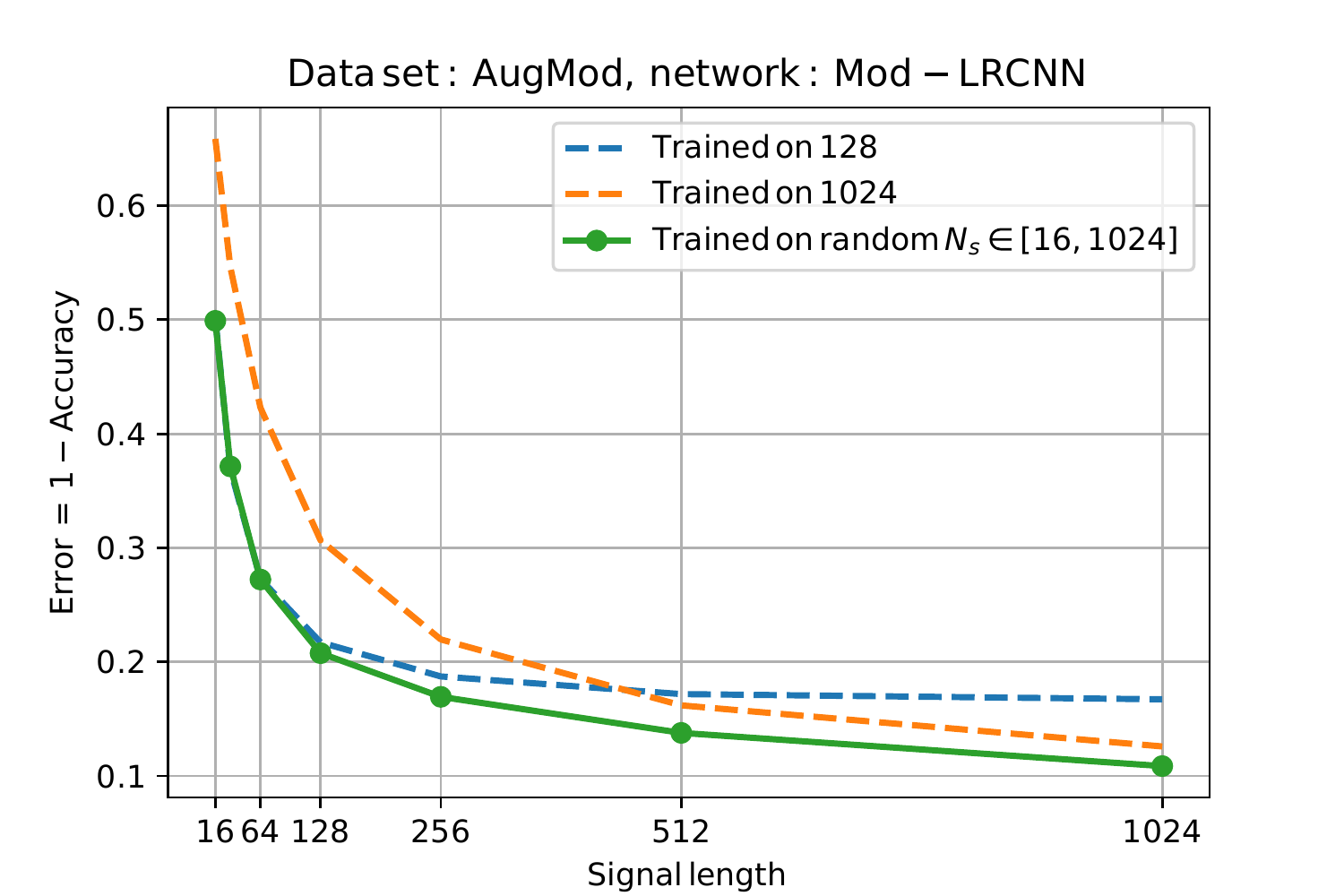}
    \caption{\textbf{Left:} Error rate as a function of the signal-to-noise ratio
    for the five different networks.
    The performances are given for the \texttt{AugMod} dataset, with
    $1024$~samples long signals.
    \textbf{Right:} Error rate of the \mbox{Mod-LRCNN} architecture,
    developed for this study, on the \texttt{AugMod}
    dataset as a function of the signal length. The blue dashed curve gives
    the performances for a model trained on $128$~samples long signals,
    the orange dashed curve for a model trained on $1024$,
    and the green unbroken curve for a training with signals of dynamically random
    sizes $N_{s} \in [16,1024]$.}
    \label{figure::SNR_signal_duration}
\end{figure*}

We compare the performances of each neural network
at classifying modulations, on the \texttt{AugMod} dataset, as a function of
signal-to-noise ratio. The results are presented on the left panel of
figure~\ref{figure::SNR_signal_duration}. The panel gives the error
rate as a function of SNR. Mod-LRCNN, developed for this study, performs more than $40$\%
better than the best architecture of previous studies, RML-ResNet, at $SNR=0$.
In the $SNR \in [0,30]$ range, Mod-LRCNN effectively improves the performances by
$\sim 5$~dB.

We assess the training time by looking at the time per epoch when running
on the \texttt{AugMod} dataset, with $1024$~samples.
Other datasets
give similar results. Mod-LRCNN runs in $3.1$~ms per example, resulting in
$27$~seconds per epoch, with $512$~examples per batch,
for a total training time of $1.5$~hours with $200$~epochs.
Mod-LCNN and RML-CNN/VGG are twice as fast, however,
RML-ResNet is $1.25$~times longer. Finally \mbox{RML-ConvNet} runs in twice as long
due to the large number of parameters (table~\ref{table::nb_parameters_networks}).
The fact that Mod-LRCNN runs each epoch in twice the time compared to
Mod-LCNN is balanced by both its higher accuracy
(table~\ref{table::perfs_accuracy}) and the fact that less epochs are needed
to converge (figure~\ref{figure::learning curves}).

\section{Specific performances of Mod-LRCNN}
\label{section::performances_of_Mod_LRCNN}

As discussed previously in section~\ref{section::comparing_the_different_architectures},
the Mod-LRCNN architecture, developed in this study, outperforms all other architectures in accuracy.
We investigate in this section its performances on different signal lengths,
and under different sets of impairments.

\subsection{Signal duration}
\label{section::signal_duration}

Mod-LCNN and Mod-LRCNN's strength are their invariance
under the signal duration. This means that once the network has been trained,
it can be used to infer the signal modulation, whatever its length.
We test this property on three different training strategies for Mod-LRCNN.
The following results are given through the implementation of Mod-LRCNN
in PyTorch \cite{paszke2019pytorch}. This choice gives us more flexibility
during training.

The right panel of figure~\ref{figure::SNR_signal_duration} presents the
classification error rate as a function of the signal length. These results are given
on the \texttt{AugMod} dataset.
The first strategy is to train Mod-LRCNN on $128$~samples long signals.
The second strategy is to train on $1024$~samples long signals.
On the test set, we limit each example to the first $\{16,32,64,128,256,512,1024\}$
samples, infer the modulation class, and
give the resulting error rate.

In the right panel of figure~\ref{figure::SNR_signal_duration}
the blue dashed curve presents the results for the first strategy,
and the orange for the second.
One could have expected \mbox{Mod-LRCNN} trained on $1024$ to outperform
the first strategy on the full range. It is the case for signals more than
$256$~samples long, however it is not the case bellow.
This indicates a tendency of Mod-LRCNN, trained on $1024$, to overfit
long signals.

We develop a third strategy where we modify
dynamically the length of the signal during training. At each batch iteration
we randomly pick an integer $N_{s} \in [16,1024]$, and limit the signal duration
to the first $N_{s}$ samples. The resulting accuracy on the
test set is given in the green unbroken curve.
We observe that indeed this new training scheme allows to get good
performances for short and long signals.

\subsection{Frequency shift}
\label{section::frequency_shift}

\begin{figure}[ht]
    \centering
    \includegraphics[width=0.99\linewidth]{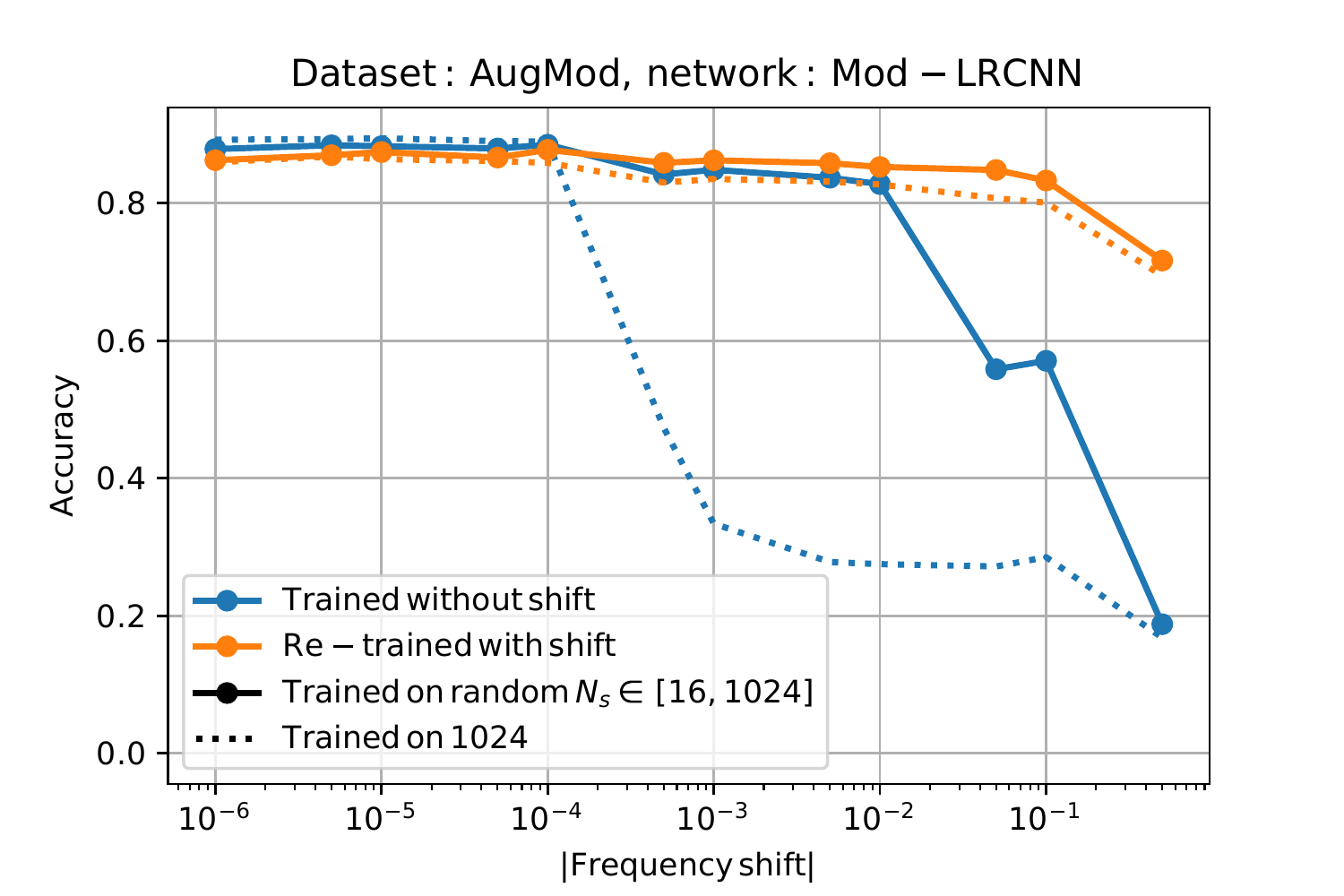}
    \caption{Accuracy of the Mod-LRCNN architecture, developed for this study,
    on the \texttt{AugMod} dataset, with $1024$ samples long signals,
    enhanced with frequency shift impairments:
    in blue the results for the network trained on a dataset without carrier
    shift, and in orange for a re-training on the dataset including it.
    Unbroken curves are for a training with variable random signal lengths,
    $N_{s} \in [16,1024]$, and
    dotted curves for training on fixed $1024$~samples long signals.}
    \label{figure::effect_frequency_offset}
\end{figure}

The \texttt{AugMod} synthetic dataset is reproduced adding a relative
frequency offset (table~\ref{table::set_of_impairments}) on top of the
other baseline impairments. We span a wide range of values, from
positive and negative $50$\% of the carrier frequency, with a logarithmic
scale: $\Delta f \in \pm [10^{-6}, 5\times10^{-1}]$.
The effect of this latter impairment is to drift the constellations into
circular patterns with a typical time scale $1/\Delta f$.
The results are presented on the left panel of 
figure~\ref{figure::effect_frequency_offset}, for Mod-LRCNN
trained on the \texttt{AugMod} dataset, with $1024$~samples long signals.

In this figure, the unbroken blue curve gives the result when Mod-LRCNN is trained
on the \texttt{AugMod} dataset without the frequency shift impairment,
with variable length of signals (sec.~\ref{section::signal_duration}).
This curve thus displays the ability of the network to generalize to out of
distribution example signals. 
The dotted blue curve presents the same
results, but for a training with fixed $1024$~samples long signals.
We observe that the accuracy starts to drop at $|\Delta f| = 10^{-4}$
and falls out at $10^{-2}$. This behavior is even more drastic
when the network is trained on fix $1024$~samples long signals
(dashed blue curve).
This later behavior confirms the tendency of networks trained on fixed
size signals to overfit long signals and thus be less robust to time
varying impairments.

The orange curves show the accuracy on the test set when Mod-LRCNN
is trained on half of the \texttt{AugMod} dataset, impaired with frequency
shifts. We recover good performances at large frequency shifts.
Following the methods of curriculum learning \cite{10.1145/1553374.1553380},
only few epochs are needed to perform this re-training, if the
weights are initialized to the best values found when trained on the
simpler \texttt{Augmod} dataset.

\section{Conclusion}
\label{section:conclusion}

This study presented an artificial neural network architecture
allowing to classify modulations: the light residual
convolutional neural network for modulation classification, ``Mod-LRCNN''.
This architecture is lighter than previously published networks.
Its architecture is invariant to the signal length, allowing it to
adapt perfectly to signals
recorded on more or less samples, without a need for re-training.
The network is designed to search for the natural symmetries of the signals,
extract latent features and use them to classify modulations.
It simply builds statistical significance with the signal duration, and thus
can process data stream.

It performs better than three public networks
\cite{10.1007/978-3-319-44188-7_16,2018ISTSP..12..168O} on all four
publicly available datasets, e.g. \texttt{RadioML2018.01A},
and on a custom made dataset, \texttt{AugMod}. It is defined by
up to two orders of magnitude less parameters.
In the $\mathrm{SNR} \in [0,30]$ range, Mod-LRCNN effectively improves the threshold by
$\sim 5$~dB (up to $10$~dB) compared to previously published networks.

We characterize some of the performances of the network. When trained on
dynamically changing examples lengths, between $16$ to $1024$~samples,
the network is able to give very good accuracy whatever the inferred
signal lengths. This training technique prevents overfitting long signals,
and thus gives good performances on evolving impairments, e.g. frequency
shift.
We show the ability of the network to efficiently classify signals
under frequency shift impairment,
even when they are out of the distribution given in the training set.
Even better performances can be obtained
through curriculum learning, by training the network in few epochs,
if the weights are initialized at their values for the simpler dataset.

The datasets introduced in this study have allowed us to train our network
to create signal representation invariant to real life impairments.
We aim at adding more complexity to this set, e.g. non-linear modulations,
multi-path propagation, and test the network
under more real indoor and outdoor propagated signals.
In another direction, we aim at looking into reducing the
complexity of the network through pruning and weight quantization, this would
allow faster and lighter real time processing of radio signals.

\bibliographystyle{IEEEtran}
\bibliography{conference_101719}

\end{document}